\documentclass{article}

\PassOptionsToPackage{authoryear}{natbib}
\usepackage[preprint]{neurips_data_2021}

\usepackage[utf8]{inputenc} 
\usepackage[T1]{fontenc}    
\usepackage{hyperref}       
\usepackage{url}            
\usepackage{booktabs}       
\usepackage{amsfonts}       
\usepackage{nicefrac}       
\usepackage{microtype}      
\usepackage{xcolor}         
\usepackage{graphicx}
\usepackage{multirow}
\usepackage{subfig}

\title{HeadlineCause: A Dataset of News Headlines for Detecting Causalities}

\author{%
  Ilya Gusev \\
  Moscow Institute of Physics and Technology\\
  Moscow, Russia \\
  \texttt{ilya.gusev@phystech.edu} \\
  \And
  Alexey Tikhonov \\
  Yandex  \\
  Berlin, Germany \\
  \texttt{altsoph@gmail.com} \\
}

\begin{document}

\maketitle

\begin{abstract}
    Detecting implicit causal relations in texts is a task that requires both common sense and world knowledge. Existing datasets are focused either on commonsense causal reasoning or explicit causal relations. In this work, we present HeadlineCause, a dataset for detecting implicit causal relations between pairs of news headlines. The dataset includes over 5000 headline pairs from English news and over 9000 headline pairs from Russian news labeled through crowdsourcing. The pairs vary from totally unrelated or belonging to the same general topic to the ones including causation and refutation relations. We also present a set of models and experiments that demonstrates the dataset validity, including a multilingual XLM-RoBERTa based model for causality detection and a GPT-2 based model for possible effects prediction.
\end{abstract}

\section{Introduction} \label{introduction}

Causality is a crucial concept in many human activities. Automatic inference of causal relations from texts is vital for any model attempting to analyze documents or predict future events. It is much more essential when we consider news. On the other hand, it is still a challenging task for text understanding models, as it requires both common sense and world knowledge.

From the practical sense, news aggregators are interested in detecting causal relations. Firstly, they should understand what news documents refute others to build a relevant and rapid news feed, and a refutation is a special kind of causal relation. For example, almost every big accident has many death toll changes that refute each other. The rise of fake news also increases the need to detect refutations. Secondly, news aggregators should be able to differentiate between news from different sources about the same event and news on the same topic but about different events. Cause-effect event pairs are hard negative samples for this task.

There are several types of causal relations. In this work, we focus on implicit inter-sentence causal relations. It means that cause and effect are in different sentences, even in different texts in our case, and there are no explicit linking words between them.

This paper introduces a dataset of news headline pairs in English and Russian with causality labels obtained through crowdsourcing. We deliberately chose not to include texts of news documents in this dataset as almost every headline contains only one fact and roughly corresponds to a notion of an event. Furthermore, using headlines is much easier than detecting causalities between different parts of texts. For the same reasons, headlines were used in other works~\citep{radinsky_prediction}.

Natural language understanding benchmarks such as SuperGLUE~\citep{wang2019superglue} and RussianSuperGLUE~\citep{shavrina2020russiansuperglue} were introduced recently and are a great way to track natural language research progress. These NLU benchmarks have inspired this work. The only task dealing with causality in these benchmarks is COPA~\citep{roemmele2011choice}  (PARus in the Russian version). The examples in this task are from the general domain and do not fully represent causal relations in other domains. Moreover, COPA was deliberately built as a benchmark but not a dataset for model training. 

The motivation of this work was a desire to know how modern models can handle implicit causal relations. Ultimately we would like to predict for every event what other events led to it and to predict possible future events.

To prove our dataset useful, we analyzed its contents, trained several BERT-family classifiers to detect causalities in previously unseen headlines, and checked their performance. We also trained GPT-2 based models to predict future headlines based on the current ones.

The resulting dataset is one of the few datasets on implicit inter-sentence causal relations. Using world knowledge and common sense is the only way to infer a causal relation for many samples. Embedding that knowledge into the models is the main challenge the dataset poses for a research community.

As for potential negative social impact, we do not see any direct malicious applications of our work.

The data probably do not contain offensive content, as news agencies usually do not produce it, and a keyword search returned nothing. Still there are news documents in the dataset on several topics some people can consider sensitive, such as deaths or crimes.

\section{Related work}

In recent years several reviews of causality extraction papers appeared~\citep{review0, review1, review2}. We will not mention all the works covered there again, but we will look at the most significant ones.

Most of the papers are describing methods to extract explicit causal relations. For example, these relations can be collected with a set of linguistic patterns  ~\citep{khoo1998automatic, khoo2, girju_text_mining} or with machine learning methods, such as decision trees~\citep{girju2003automatic} or SVM~\citep{bethard-martin-2008-learning} over syntactic and semantic features. ~\citet{riaz-girju-2013-toward} explore causal associations of verb-verb pairs for this purpose.

As for effect prediction, there is a work of~\citet{radinsky_prediction} about a prediction of future events through building a generalizing abstraction tree over given event pairs. A new event is matched to a node of this tree, and an associated prediction rule is applied to produce effects. The authors obtain the event pairs from news headlines, but, in contrast with our work, cause and effect should be in the same headline.

The Topic Detection and Tracking initiative (TDT)~\citep{allan1998topic} and its successors is a related research area we should mention. The area is mainly about the detection of news events, topics, and composing storylines. However, the already clustered news collection makes composing causality graphs or predicting new events much more manageable.

\citet{radinsky_2} use TDT methods to compose storylines through text clustering. Then they use these storylines as a heuristic for identifying possible causal relationships among events. Over these storylines, they are trying to predict the probabilities of various future events. This work is very close to ours, as the causal relations in this work are implicit, and we also utilize text clustering methods as one of the heuristics for sampling candidate pairs.

The field of event evolution~\citep{event_evolution_graphs,event_evolution_social_streams} is the other TDT successor. The event evolution graph built in~\citet{event_evolution_graphs} can also be seen as a causality graph.

There were also attempts to build causality datasets. The Event StoryLine Corpus~\citep{esc} is one of these attempts focusing on complex annotations of the events and links between them. The modification by \citet{caselli-inel-2018-crowdsourcing} involves mining causal relations through crowdsourcing to enhance the dataset. The other dataset, Altlex~\citep{altlex}, leverages parallel Wikipedia corpora to identify
new causality markers.

Several recent papers focus on implicit inter-sentence causal relations, including~\citet{Jin2020IntersentenceAI, hosseini2021predicting}. ~\citet{Jin2020IntersentenceAI} focus on models for extracting these relations from Chinese corpora, and ~\citet{hosseini2021predicting} use BERT and its modifications to detect the directionality of these relations.

There is also a paper by \citet{Laban2021NewsHG} which focuses on converting an event detection task to a classical NLU task on headlines. The main idea and methodology of this work are very similar to ours, but the target class differs. Our main goal is to predict headlines with causal relations, and in contrast, their goal is to predict headlines about the same event.

\section{Data}

\subsection{Definitions}

\textbf{Same headlines}: two headlines are considered as the same if they are about the same things or differ in minor details. In other words, if they describe the same event, they should be considered the same.

An example of headlines we consider same (sample en\_tg\_572):\\
A: \textit{Exclusive: NextVR acquired by Apple (Updated)}\\
B: \textit{Apple Buys Virtual Reality Company NextVR}

\textbf{Causality}: the first headline causes the second headline if the second headline is impossible without the first one. If the first event did not happen, then the second event must not be happening too.

An example of a news headline pair with a causal relation (sample en\_tg\_1153):\\
A: \textit{Oklahoma spent \$2 million on malaria drug touted by Trump}\\
B: \textit{Gov. Kevin Stitt defends \$2 million purchase of malaria drug touted by Trump}

This type of causality is known as necessity causality. There are other possible definitions of causality, including sufficient causality or a cost-based concept of causality. They are described in~\citet{roemmele2011choice}. We have several reasons to use this particular definition. First, it is aligned with our goals stated in the introduction. Second, it is easy enough to be used in a crowdsourcing project.

\textbf{Refutation}: the second headline refutes the first one if the second headline makes the first one irrelevant. Refutations are a subset of causal pairs, and every refuting headline is an effect of some cause, but not every effect of a cause is a refutation.

An example of a news headline pair with a refutation (sample en\_tg\_496):\\
A: \textit{Report: Microsoft acquiring Microvision, a leader in ultra-miniature projection display}\\
B: \textit{Microsoft denies MicroVision acquisition}

\subsection{Sources} \label{sources}
We used two sources of news documents: the Lenta corpus\footnote{\url{https://github.com/yutkin/Lenta.Ru-News-Dataset}} and documents from the Telegram Data Clustering Contest\footnote{\url{https://contest.com/docs/data\_clustering2}}. We additionally parsed the Lenta website\footnote{\url{https://lenta.ru}} to obtain fresher news documents and hyperlinks between documents. Lenta is one of the oldest Russian news websites, and the dataset contains over 800 thousand news from 1999 to 2020. The Telegram news dataset was published in 2020 and contains data from hundreds of sources in over ten languages from October 2019 to May 2020.

The original data for the Telegram dataset contains various types of documents. We trained a FastText classifier on separate crowdsourcing annotations and open datasets to differentiate news from other documents. These annotations and the classifier itself are available in the separate GitHub repository~\footnote{\url{https://github.com/IlyaGusev/tgcontest}}. The classifier is not ideal, but the majority of the resulting documents are news.

The licenses for both datasets are not provided. Our dataset uses only news headlines, provides links for all used documents, and provides authors where possible, which can be considered fair use. We also emailed Lenta asking for permission to publish their data.

\subsection{Candidates sampling} \label{candidates}
For both datasets, we used four filters to extract candidates for annotation:
\begin{itemize}
    \item A presence of a hyperlink between two documents
    \item An affiliation of documents to the same website
    \item A cosine distance between LaBSE embeddings~\citep{feng2020languageagnostic} with a threshold
    \item A presence of different locations in headlines
\end{itemize}
We combined these filters in various combinations in different annotation pools to collect diverse data. We did not have a specific algorithm or scheme for applying these filters and used some of them as different problems emerged. For example, at some moment we observed that our trained model was making errors in linking news headlines about similar events, but in different locations, so we annotated more data with the last filter. The first two filters were used as the main ones.

Another problem we detected was a subset of headlines that our models considered causes without even looking at effects. This type of error is common for sentence-pair inference tasks~\citep{Gururangan2018AnnotationAI}. To handle this, we trained a single headline classifier to estimate a priori probability of a headline being cause or effect, sampled negative pairs for the original task from the strongest examples, and annotated them.

\begin{figure}
    \centering
    \includegraphics[width=1.0\linewidth]{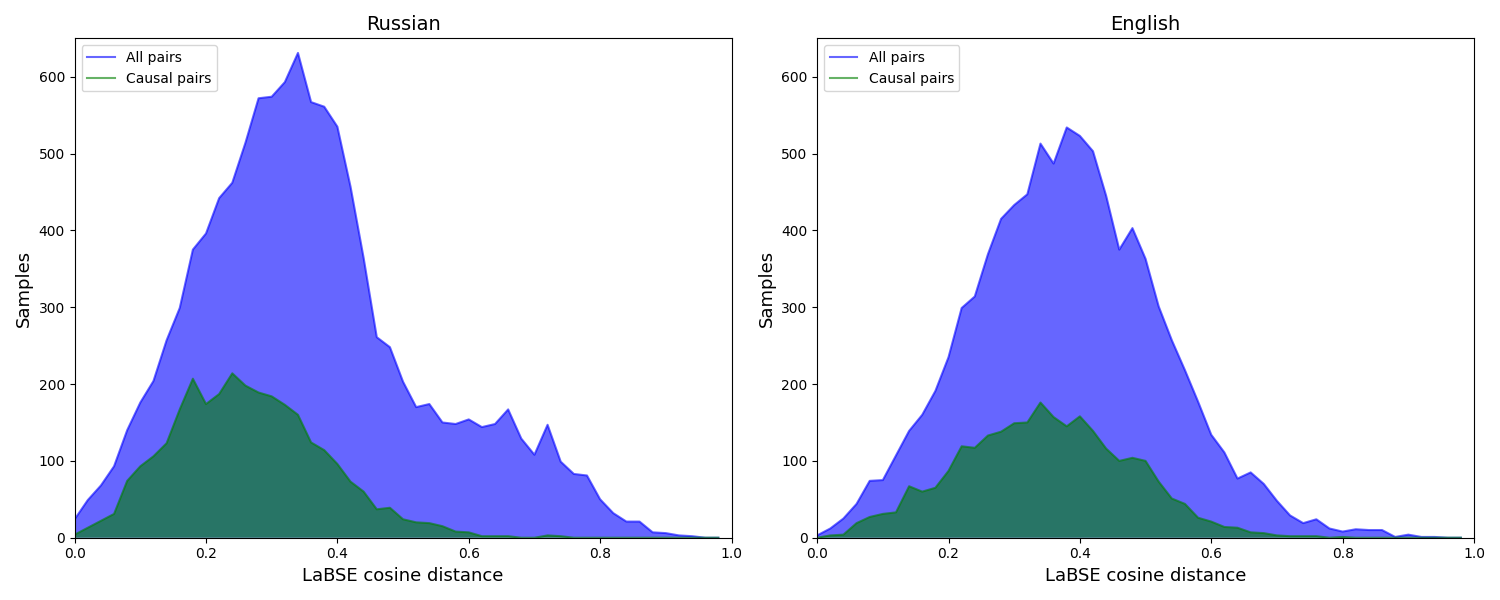}
    \caption{A dependency between a LaBSE distance and the number of causal pairs}
    \label{fig:labse2}
\end{figure}

\begin{figure}
    \centering
    \includegraphics[width=1.0\linewidth]{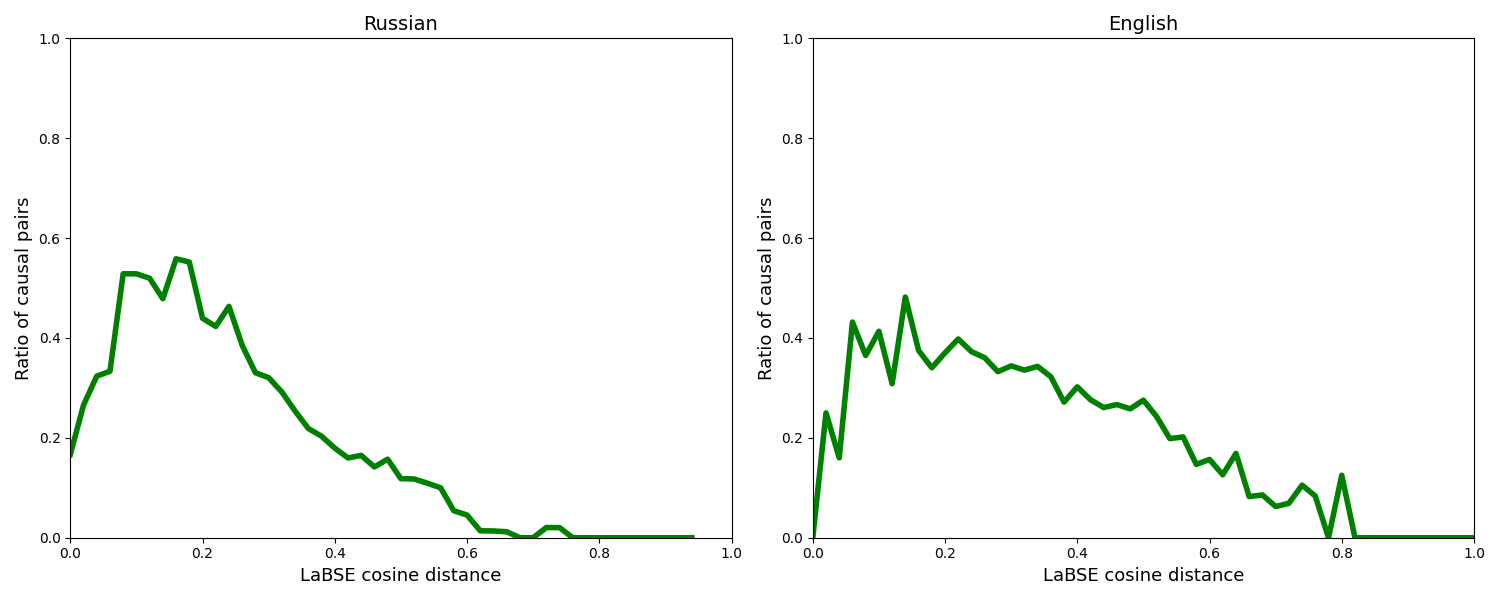}
    \caption{A dependency between a LaBSE distance and the ratio of causal pairs}
    \label{fig:labse}
\end{figure}

To estimate a final role of a sampling method based on a cosine distance between LaBSE embeddings, we plotted a dependency between the distance and the number of causal pairs. It can be seen in Figure~\ref{fig:labse}. We used this distance as an upper threshold in our candidate sampling. It is clear from the figure that for the Russian annotations we did not lose any of the information as with the increasing threshold the number of causal pairs is decreasing very fast. It is not so obvious for the English annotations, so we additionally plotted Figure~\ref{fig:labse2}. One can see that the ratio of causal pairs is also decreasing.

To estimate a final role of a sampling method based on a cosine distance between LaBSE embeddings, we plotted a dependency between the distance and the number of causal pairs. It can be seen in Figure~\ref{fig:labse}. We used this distance as an upper threshold in several candidate samplings. It is clear from the figure that we did not lose many causal pairs above the threshold for the Russian annotations, as with the increasing threshold, the number of causal pairs is decreasing very fast. For English annotations, this regularity is not apparent from Figure~\ref{fig:labse}, so we additionally plotted Figure~\ref{fig:labse2}. One can see that the ratio of causal pairs in English is also decreasing with the distance.

\subsection{Annotation} \label{annotation}

\begin{figure}
    \centering
    \includegraphics[width=1.0\linewidth]{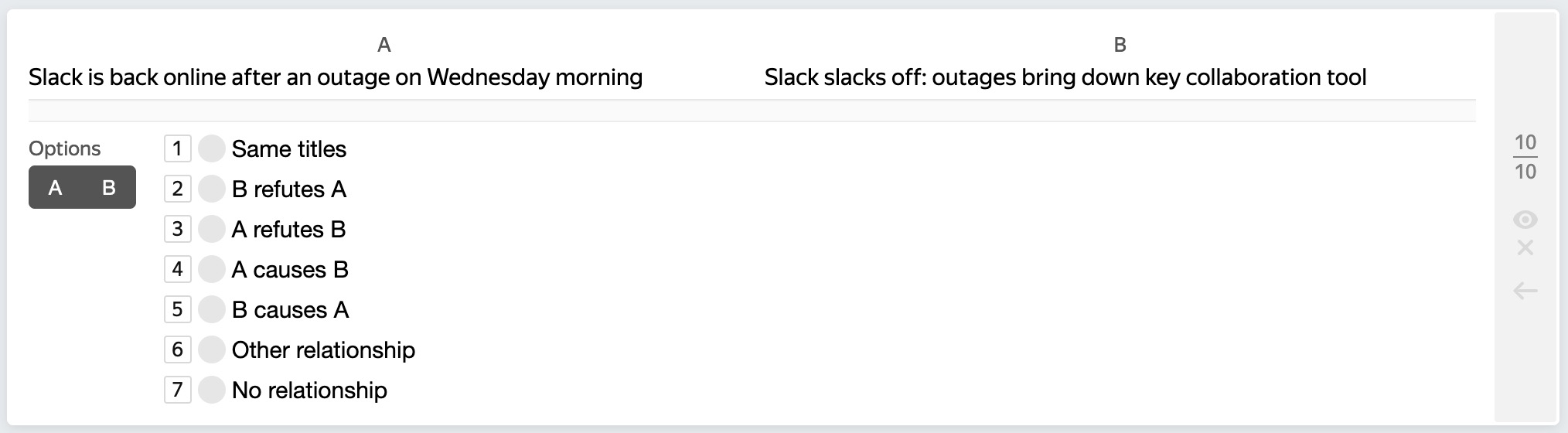}
    \caption{English annotation interface for desktop devices}
    \label{fig:annotation_interface}
\end{figure}

We annotated every candidate pair with Yandex Toloka\footnote{\url{https://toloka.ai/}}, a crowdsourcing platform. We chose this platform as we are very familiar with it, and it has many workers that are native Russian speakers, which is useful for Russian annotation. The task was to determine a relationship between two headlines, A and B. There were seven possible options: titles are almost the same, A causes B, B causes A, A refutes B, B refutes A, A linked with B in another way, A is not linked to B. An annotation guideline was in Russian\footnote{\url{https://ilyagusev.github.io/HeadlineCause/toloka/ru/instruction.html}} for Russian news and in English\footnote{\url{https://ilyagusev.github.io/HeadlineCause/toloka/en/instruction.html}} for English news. Ten workers annotated every pair. The annotation interface is in Figure~\ref{fig:annotation_interface}. The total annotation budget was 2173\$, with the estimated hourly wage paid to participants of 1.13\$. Initially, it was 45 cents, but we reconsidered this wage because of ethical reasons, as it is lower than the minimum wage in Russia. Annotation management was semi-automatic. Scripts are available in the project repository\footnote{\url{https://github.com/IlyaGusev/HeadlineCause}}.

As for the quality control, we required annotators to pass training, exam, and their work was continuously evaluated through the control pairs ("honeypots"). The threshold was 70\% correct examples for training and 80\% correct examples for exams and honeypots. No additional language abilities check was done, as these thresholds should remove workers without them. All examples from training and exam are also available in the project repository.

\begin{table}
  \caption{Annotation statistics}
  \label{annot_stats}
  \centering
  \subfloat[Overall numbers]{\begin{tabular}{lll}
    \toprule
    & English & Russian \\
    \midrule
    Total number of pairs & 10078 & 11649 \\
    Number of pairs with links & 8737 & 5241 \\
    Number of pairs from the same source & 8139 & 8278 \\
    Number of workers & 180 & 457 \\
    Average number of tasks per worker & 560 & 255 \\
    Total budget & 1008\$ & 1165\$ \\
    \bottomrule
  \end{tabular}}
  \quad
  \subfloat[English proj., top-6 countries]{\begin{tabular}{ll}
    \toprule
    Country & Workers \\
    \midrule
    India & 26 \\
    Kenya & 24 \\
    The Philippines & 19  \\
    Turkey & 10 \\
    Nigeria & 9 \\
    Pakistan & 7 \\
    \bottomrule
  \end{tabular}}
\end{table}

Annotation statistics are in Table~\ref{annot_stats}. As for the number of pairs with links and the same sources, the filtering system causes these high numbers. As for the number of annotators, historically, the Russian annotation was done earlier than English and was split into two considerable periods, so more workers were involved.

\subsection{Aggregation} \label{aggregation}

\begin{table}
  \caption{Task labels alignment}
  \label{task-alignment}
  \centering
  \begin{tabular}{ll}
    \toprule
    \textbf{Full} & \textbf{Simple} \\
    \midrule
    Left-right causality & \multirow{2}{8em}{Left-right causality} \\
    Left-right refutation &  \\
    \midrule
    Right-left causality & \multirow{2}{8em}{Right-left causality} \\
    Right-left refutation &  \\
    \midrule
    Same event & \multirow{3}{8em}{No causality} \\
    Other relationship &  \\
    No relationship & \\
    \bottomrule
  \end{tabular}
\end{table}

We aggregate annotations in two settings. The first setting, Full, includes all seven possible classes. The second setting, Simple, unites some of the classes to simplify the task. The alignment between labels is presented in Table~\ref{task-alignment}. We include only samples with an agreement of 70\% or higher in the final dataset. The agreement is calculated relative to the setting, with three labels for the Simple setting and seven labels for the Full setting.

\begin{table}
  \caption{Agreement distribution for both languages and both settings, every sample was annotated by ten people, $\alpha$ is the Krippendorff's alpha~\citep{krippendorff2011computing}, computed with NLTK package~\citep{nltk}}
  \label{agreement}
  \centering
  \begin{tabular}{lllllllll}
    \toprule
    & English, Simple & English, Full & Russian, Simple & Russian, Full \\
    \midrule
    10 votes & 966 (10\%) & 167 (2\%) & 5379 (46\%) & 2783 (24\%) \\
    9 votes & 1476 (14\%) & 450 (4\%) & 2058 (18\%) & 1532 (13\%) \\
    8 votes & 1568 (16\%) & 856 (8\%) & 1384 (12\%) & 1497 (13\%) \\
    7 votes & 1703 (17\%) & 1227 (12\%) & 1059 (9\%) & 1603 (14\%) \\
    6 votes & 1905 (19\%) & 1734 (17\%) & 983 (8\%) & 1770 (15\%) \\
    5 votes & 1784 (18\%) & 2309 (23\%) & 683 (6\%) & 1597 (14\%) \\
    4 votes & 676 (7\%) & 2151 (22\%) & 103 (1\%) & 717 (6\%) \\
    3 votes & 0 (0\%) & 1127 (11\%) & 0 (0\%) & 150 (1\%) \\
    2 votes & 0 (0\%) & 57 (1\%) & 0 (0\%) & 0 (0\%)\\
    \midrule
    Total & 10078 & 10078 & 11649 & 11649 \\
    Average agreement & 0.699 & 0.548 & 0.862 & 0.745 \\
    $\alpha$, all samples & 0.289 & 0.255 & 0.598 & 0.548 \\
    $\alpha$, 7 or more votes & 0.458 & 0.551 & 0.708 & 0.733 \\
    \bottomrule
  \end{tabular}
\end{table}

Agreement distribution for both tasks is in Table~\ref{agreement}. Agreement between workers is very different for Russian and English. There are several possible reasons for this. Firstly, English workers are less homogeneous than Russian ones, as they are from a more extensive list of countries, and English is not native for some of them. Secondly, there is a difference in the complexity of the task itself. The headlines are more challenging in English documents, as they include a bigger list of entities, including local ones. The diversity of news agencies is also more considerable in the English dataset than in Russian.

The distribution by the most popular countries of English-speaking workers is in Table~\ref{annot_stats}. Most of the workers are probably not native speakers. It affects the annotation quality, but the aggregation with an overlap of 10 and tight quality control should help to maintain it.

We use the Majority vote (MV) aggregation method. There are several possible alternatives: the~\citet{dawid_skene} method, aggregation by skill, and some others. Some of them are supported by crowdsourcing platform itself. We chose the MV as it is easily interpretable and yielded consistent results in our experiments. We also tried to use the Dawid-Skene method, but the training on resulting annotations was hard and yielded poor metrics, so we abandoned it early. However, we still do not know whether the poor results came from the method itself or the complexity of the examples it brings, and it is the subject of future experiments.

\begin{table}
  \caption{Simple task aggregated data statistics after excluding samples with less than seven votes and additional postprocessing (Section \ref{postprocessing})}
  \label{stat_simple}
  \centering
  \begin{tabular}{lllll}
    \toprule
    & English & Russian \\
    \midrule
    Left-right causality & 720 (13\%) & 1173 (12\%) \\
    Right-left causality & 610 (11\%) & 1224 (13\%) \\
    No causality & 4086 (76\%) & 7156 (75\%) \\
    \midrule
    Total & 5416 & 9553 \\
    \bottomrule
  \end{tabular}
\end{table}

The final statistics for aggregated annotations are presented in Table~\ref{stat_simple} and Table~\ref{stat_full}. All datasets are imbalanced. For the Simple task, only 25\% of samples contain causal relations.

\begin{table}
  \caption{Full task aggregated data statistics after excluding samples with less than seven votes and additional postprocessing (Section \ref{postprocessing})}
  \label{stat_full}
  \centering
  \begin{tabular}{lllll}
    \toprule
    & English & Russian \\
    \midrule
    Left-right causality & 428 (17\%) & 914 (13\%) \\
    Right-left causality & 386 (15\%) & 966 (13\%) \\
    Left-right refutation & 61 (2\%) & 126 (2\%) \\
    Right-left refutation & 34 (1\%) & 127 (2\%) \\
    Same event & 254 (10\%) & 780 (11\%) \\
    Other relationship & 813 (32\%) & 1655 (23\%) \\
    No relationship & 536 (21\%) & 2575 (36\%) \\
    \midrule
    Total & 2512 & 7143 \\
    \bottomrule
  \end{tabular}
\end{table}

\subsection{Postprocessing} \label{postprocessing}

We remove pairs that are not consistent with timestamps. In other words, if a temporally following headline is a cause of a temporally preceding headline, we consider an annotation of this pair inaccurate. There were 981 (10\%) such pairs for the English dataset and 540 (5\%) pairs for the Russian one.

These numbers drop to 202 (5\%) and 132 (2\%) if we consider only pairs with more than 70\% agreement (Simple task) and from the same sources. Different sources can have different timestamps policies or different promptness of reaction to the events, so it can be incorrect to compare them.

\subsection{Splits}
We split the dataset into train, validation, and test sets by time. The main reason is a possible entity and event bias in the news domain. Trained models should work well with the unseen entities, locations, and events, and the best way to emulate these factors is to split by time. We consider a maximum of left and right timestamps as a timestamp of a pair. The training dataset contains the first 80\% of pairs, the validation dataset contains the next 10\% of pairs, and the test dataset utilizes the remaining 10\%.

Lenta and Telegram corpora have very different densities of news in time and different time spans affected. So we split these two datasets separately and unite the resulting splits from both sources, so we have samples from both datasets in the train, validation, and test sets equally presented.




\subsection{Augmentations}
We apply two augmentations\footnote{\url{https://github.com/IlyaGusev/HeadlineCause/blob/main/headline_cause/augment.py}} to the train and validation datasets. The first one adds symmetrical pairs to enforce a model to be logical in case of swapping the headlines. The second one adds typos to the left, right, or both headlines to make a model more robust.

The symmetrical augmentation doubles the size of the dataset. Typos are applied for 5\% of the dataset, with original pairs preserved. Typos are just swaps of adjacent letters.

\section{Experiments} \label{experiments}

To prove our dataset to be helpful, we trained several models for the Simple and Full tasks. The main models use XLM-RoBERTa-large~\citep{conneau-etal-2020-unsupervised} as a base pretrained model. It is multilingual and allows Russian inputs and English ones, still providing a good classification quality.

The training was done using GPU at the Google Colab Pro platform. The code is publicly available. One entire training run for one task requires at most 120 minutes. We trained and evaluated every model three times with different random seeds.

The exact hyperparameters can be found in the training notebook itself. They are standard for BERT-family models.

\subsection{Simple task}

For this task, we consider causality ROC AUC on two classes as a main metric. To calculate it, we unite Left-right and Right-left classes to be able to vary a classifier threshold.

We used a CatBoost~\citep{catboost} classifier over TF-IDF features as a baseline model. The primary reason to use such a model is to check lexical biases or data leaks. For this task, causality ROC AUC was 71\% for Russian and 62\% for English, so we concluded no huge leaks.

The results for the Simple task are in Table~\ref{simple_en_ru_results}. Models are working better with Russian as the training dataset for Russian is more extensive, and the average agreement on the remaining samples is higher than for English. The final score for both languages is over 95\%, which means the models can predict test set labels far from random.

\begin{table}
  \caption{Simple task EN+RU XLM-RoBERTa results on the test sets, 3 runs}
  \label{simple_en_ru_results}
  \centering
  \begin{tabular}{lllll}
    \toprule
    & \multicolumn{2}{c}{English} & \multicolumn{2}{c}{Russian} \\
    & Samples & Score, \% & Samples & Score, \% \\
    \midrule
    No causality F1 & 421 (78\%) & 94.1 $\pm$ 0.2 & 782 (82\%) & 94.7 $\pm$ 0.4 \\
    Left-right F1 & 65 (12\%) & 75.2 $\pm$ 1.4 & 99 (10\%) & 76.7 $\pm$ 2.0 \\
    Right-left F1 & 56 (10\%) & 70.0 $\pm$ 1.5 & 76 (8\%) & 69.9 $\pm$ 2.0  \\
    \midrule
    Accuracy & 542  & 89.4 $\pm$ 0.2 & 957 & 90.9 $\pm$ 0.7  \\
    Causality ROC AUC & 542 & 96.3 $\pm$ 0.2  & 957 & 95.6 $\pm$ 0.2 \\
    \bottomrule
  \end{tabular}
\end{table}

We use a checklist \citep{checklist} methodology to evaluate different aspects of the model. We do not include in Table~\ref{simple_en_ru_checklist} tests that models pass without fail, only those that fail in a considerable number of cases. The typos test ensures the robustness of models, swapping order checks their logic, and the different locations test inspects whether they can detach causes and effects that are certainly not connected. The swap-order test is also similar to Commutative Test from \citet{Laban2021NewsHG}. The first two groups of tests match the augmentation methods introduced to reduce the failure rate. 

\begin{table}
  \caption{Simple task EN+RU XLM-RoBERTa checklist results, best model}
  \label{simple_en_ru_checklist}
  \centering
  \begin{tabular}{lll}
    \toprule
    Test type and description & English, failure rate & Russian, failure rate \\
    \midrule
    INV: Adding typos & 3.5\% & 2.9\% \\
    INV: Swapping order of not causal pairs &  2.8\% & 2.0\% \\
    DIR: Swapping order of causal pairs  &  22.0\% & 12.2\%\\
    MFT: Explicit refutations with different locations & 9.5\% & 2.9\% \\
    \bottomrule
  \end{tabular}
\end{table}

\begin{figure}
    \centering
    \includegraphics[width=1.0\linewidth]{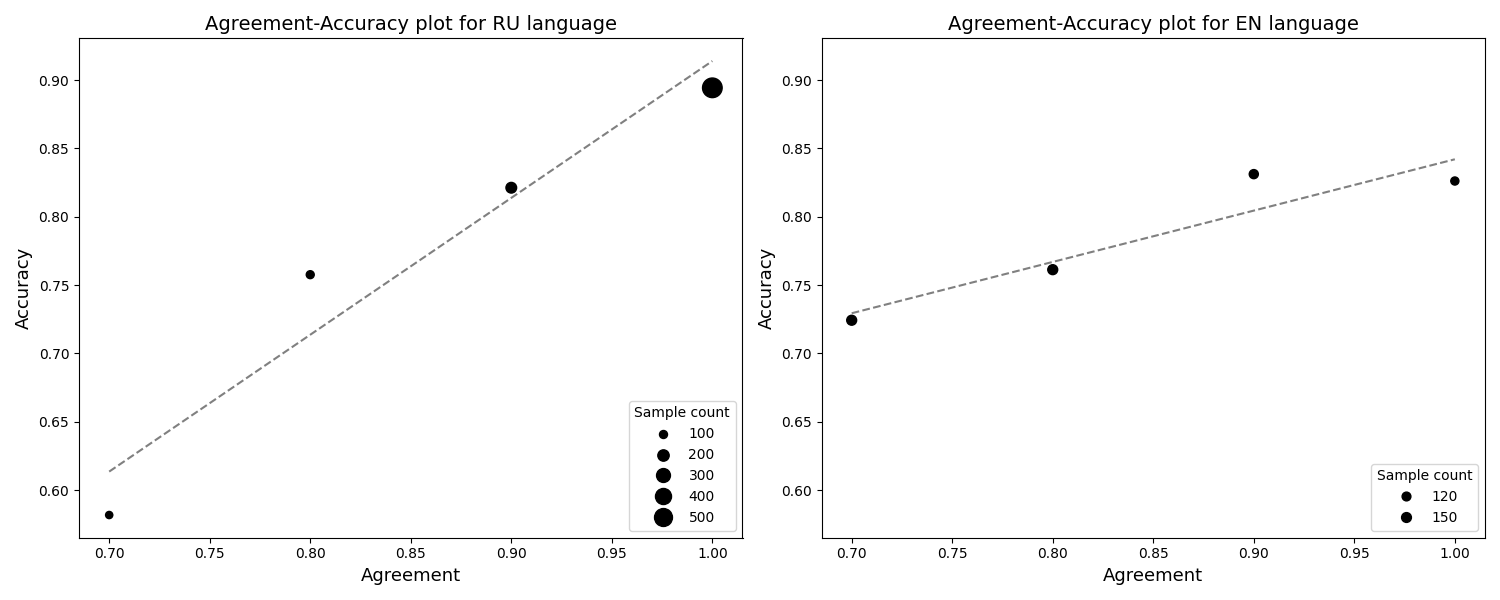}
    \caption{The dependence of the accuracy of the Simple model on annotators' agreement}
    \label{fig:accuracy_agreement}
\end{figure}

We also hypothesized that the accuracy of models is higher for samples with a higher agreement. Indeed, as shown in Figure~\ref{fig:accuracy_agreement}, it is an almost linear dependence between these two parameters, so we conclude that the agreement can be used as a proxy value of the complexity of a specific pair.

\subsection{Full task}
The results for this task are in Table~\ref{full_en_ru_results}. We provide F-score for every class and a total multiclass accuracy. The number of samples with refutation is too small to define whether we can reliably detect refutations with our model.

\begin{table}
  \caption{Full task EN+RU XLM-RoBERTa results on the test sets, 3 runs}
  \label{full_en_ru_results}
  \centering
  \begin{tabular}{lllll}
    \toprule
    & \multicolumn{2}{c}{English} & \multicolumn{2}{c}{Russian} \\
     & Samples & Score, \% & Samples & Score, \% \\
    \midrule
    No relationship F1 & 68 (27\%) & 87.7 $\pm$ 1.8 & 315 (44\%) & 95.7 $\pm$ 0.3 \\
    Same event F1 & 22 (9\%) & 81.2 $\pm$ 5.2 & 71 (10\%) & 90.7 $\pm$ 1.9 \\
    Other relationship F1 & 67 (27\%) & 80.5 $\pm$ 1.5 & 162 (22\%) & 81.3 $\pm$ 1.6 \\
    Left-right causality F1 & 43 (17\%) & 94.3 $\pm$ 0.9 & 77 (11\%) & 84.7 $\pm$ 1.1 \\
    Right-left causality F1 & 39 (15\%) & 84.1 $\pm$ 0.9 & 58 (8\%) & 77.1 $\pm$ 1.9 \\
    Left-right refutation F1 & 5 (2\%) & 25.7 $\pm$ 23.7 & 16 (2\%) & 53.8 $\pm$ 5.4 \\
    Right-left refutation F1 & 8 (3\%) & 48.1 $\pm$ 7.0 & 16 (2\%) & 74.8 $\pm$ 1.5 \\
    \midrule
    Total number of pairs & 252 & & 715 & \\
    Accuracy & & 83.5 $\pm$ 0.2 & & 87.9 $\pm$ 0.9 \\
    \bottomrule
  \end{tabular}
\end{table}

\subsection{GPT-2}
Additionally, we trained a GPT-2~\cite{radford2019language} model to predict effect headlines for the causes. The examples of such predictions are in Table~\ref{gpt_gen}. One can see that some continuations are reasonable, and it is possible to select the probable ones. Still, many continuations are not correct, and probably the larger dataset and the larger model will fix that.

In future, the generator can be used to create an augmented dataset, as it can generate grammatically correct but meaningless continuations. It will be much harder task for a detection model to correctly identify such examples.

\begin{table}
  \caption{GPT generation examples}
  \label{gpt_gen}
  \centering
  \begin{tabular}{lllll}
    \toprule
    Armed protesters demonstrate inside Michigan state capitol\\
    \midrule
    => The Michigan capitol clashes with armed protesters.\\
    => Govt defends lockdown in Michigan state capitol.\\
    => State capitol protesters demand lockdown of state capitol, condemn lockdown.\\
    => More arrests as protesters protest state capitol.\\
    => The Michigan capitol is being shut down, protesters march in defiance of state law.\\
    \bottomrule
  \end{tabular}
\end{table}



\section{Discussion} \label{discussion}

\textbf{No span annotations.} This work is different from other papers in the field in terms of annotation format. Usually, in the extraction of causal relations, the particular verbs or noun phrases are annotated. Instead, we label a whole headline pair. The primary motivation is that classification is a much more manageable task for crowdsourcing than real relation extraction. We were also inspired by sentence-level NLU tasks with no spans.

\textbf{Unclear method of candidates sampling.} We did not develop a specific and reliable schema of sampling candidates and encountered two different problems linked with this during the annotation process. The first problem was about single-sentence bias. The second problem was about the detection of causation in headlines with different locations but similar events. Both problems are described in Section~\ref{candidates}.

\textbf{Annotation aggregation.} The Majority vote is probably not the best choice, as it could leave only simple pairs. Future experiments should determine the best method. 

\textbf{Poor refutation annotation.} The collection of refutation relations was one of the main goals of our work, and it is unreached. It is possible to collect more refutations from the same document collections. For instance, one can utilize some active learning.

\textbf{Disparity between Russian and English parts.} The English part of the dataset has a lower inter-annotator agreement for several reasons we discussed in Section~\ref{aggregation}. It also affects Figure~\ref{fig:labse} and Figure~\ref{fig:accuracy_agreement}.

\textbf{No ablation study for augmentations.} The augmentations were based on the checklist's tests, and they improve results on them, but we do not present an ablation study here.

\section{Conclusion}
This paper introduced HeadlineCause, the novel dataset for implicit inter-sentence causation detection based on news headlines in Russian and English. We described the annotation process and several possible biases that we detected and tried to avoid. The dataset differs from other datasets for causal relation extraction and is more similar to NLU datasets. We also presented baselines for this dataset. We believe that HeadlineCause can be successfully used to train causal relation detection models with the subsequent composing of causation graphs.

\bibliography{papers}

\begin{thebibliography}{32}
\providecommand{\natexlab}[1]{#1}
\providecommand{\url}[1]{\texttt{#1}}
\expandafter\ifx\csname urlstyle\endcsname\relax
  \providecommand{\doi}[1]{doi: #1}\else
  \providecommand{\doi}{doi: \begingroup \urlstyle{rm}\Url}\fi

\bibitem[Allan et~al.(1998)Allan, Carbonell, Doddington, Yamron, and
  Yang]{allan1998topic}
J.~Allan, J.~Carbonell, G.~Doddington, J.~Yamron, and Y.~Yang.
\newblock Topic detection and tracking pilot study: Final report.
\newblock In \emph{Proceedings of the DARPA Broadcast News Transcription and
  Understanding Workshop}, pages 194--218, Lansdowne, VA, USA, Feb. 1998.
\newblock 007.

\bibitem[Asghar(2016)]{review0}
N.~Asghar.
\newblock Automatic extraction of causal relations from natural language texts:
  A comprehensive survey, 2016.

\bibitem[Bethard and Martin(2008)]{bethard-martin-2008-learning}
S.~Bethard and J.~H. Martin.
\newblock Learning semantic links from a corpus of parallel temporal and causal
  relations.
\newblock In \emph{Proceedings of ACL-08: HLT, Short Papers}, pages 177--180,
  Columbus, Ohio, June 2008. Association for Computational Linguistics.
\newblock URL \url{https://aclanthology.org/P08-2045}.

\bibitem[Bird et~al.(2009)Bird, Klein, and Loper]{nltk}
S.~Bird, E.~Klein, and E.~Loper.
\newblock \emph{Natural language processing with Python: analyzing text with
  the natural language toolkit}.
\newblock " O'Reilly Media, Inc.", 2009.

\bibitem[Caselli and Inel(2018)]{caselli-inel-2018-crowdsourcing}
T.~Caselli and O.~Inel.
\newblock Crowdsourcing {S}tory{L}ines: Harnessing the crowd for causal
  relation annotation.
\newblock In \emph{Proceedings of the Workshop Events and Stories in the News
  2018}, pages 44--54, Santa Fe, New Mexico, U.S.A, Aug. 2018. Association for
  Computational Linguistics.
\newblock URL \url{https://aclanthology.org/W18-4306}.

\bibitem[Caselli and Vossen(2017)]{esc}
T.~Caselli and P.~Vossen.
\newblock The event {S}tory{L}ine corpus: A new benchmark for causal and
  temporal relation extraction.
\newblock In \emph{Proceedings of the Events and Stories in the News Workshop},
  pages 77--86, Vancouver, Canada, Aug. 2017. Association for Computational
  Linguistics.
\newblock \doi{10.18653/v1/W17-2711}.
\newblock URL \url{https://aclanthology.org/W17-2711}.

\bibitem[Conneau et~al.(2020)Conneau, Khandelwal, Goyal, Chaudhary, Wenzek,
  Guzm{\'a}n, Grave, Ott, Zettlemoyer, and
  Stoyanov]{conneau-etal-2020-unsupervised}
A.~Conneau, K.~Khandelwal, N.~Goyal, V.~Chaudhary, G.~Wenzek, F.~Guzm{\'a}n,
  E.~Grave, M.~Ott, L.~Zettlemoyer, and V.~Stoyanov.
\newblock Unsupervised cross-lingual representation learning at scale.
\newblock In \emph{Proceedings of the 58th Annual Meeting of the Association
  for Computational Linguistics}, pages 8440--8451, Online, July 2020.
  Association for Computational Linguistics.
\newblock \doi{10.18653/v1/2020.acl-main.747}.
\newblock URL \url{https://aclanthology.org/2020.acl-main.747}.

\bibitem[Dawid and Skene(1979)]{dawid_skene}
A.~P. Dawid and A.~M. Skene.
\newblock Maximum likelihood estimation of observer error-rates using the em
  algorithm.
\newblock \emph{Journal of the Royal Statistical Society. Series C (Applied
  Statistics)}, 28\penalty0 (1):\penalty0 20--28, 1979.
\newblock ISSN 00359254, 14679876.
\newblock URL \url{http://www.jstor.org/stable/2346806}.

\bibitem[Feng et~al.(2020)Feng, Yang, Cer, Arivazhagan, and
  Wang]{feng2020languageagnostic}
F.~Feng, Y.~Yang, D.~Cer, N.~Arivazhagan, and W.~Wang.
\newblock Language-agnostic bert sentence embedding, 2020.

\bibitem[Girju(2003)]{girju2003automatic}
R.~Girju.
\newblock Automatic detection of causal relations for question answering.
\newblock In \emph{Proceedings of the ACL 2003 Workshop on Multilingual
  Summarization and Question Answering - Volume 12}, MultiSumQA '03, page
  76–83, USA, 2003. Association for Computational Linguistics.
\newblock \doi{10.3115/1119312.1119322}.
\newblock URL \url{https://doi.org/10.3115/1119312.1119322}.

\bibitem[Girju and Moldovan(2002)]{girju_text_mining}
R.~Girju and D.~I. Moldovan.
\newblock Text mining for causal relations.
\newblock In \emph{Proceedings of the Fifteenth International Florida
  Artificial Intelligence Research Society Conference}, page 360–364. AAAI
  Press, 2002.
\newblock ISBN 157735141X.

\bibitem[Gururangan et~al.(2018)Gururangan, Swayamdipta, Levy, Schwartz,
  Bowman, and Smith]{Gururangan2018AnnotationAI}
S.~Gururangan, S.~Swayamdipta, O.~Levy, R.~Schwartz, S.~R. Bowman, and N.~A.
  Smith.
\newblock Annotation artifacts in natural language inference data.
\newblock In \emph{NAACL}, 2018.

\bibitem[Hidey and McKeown(2016)]{altlex}
C.~Hidey and K.~McKeown.
\newblock Identifying causal relations using parallel {W}ikipedia articles.
\newblock In \emph{Proceedings of the 54th Annual Meeting of the Association
  for Computational Linguistics (Volume 1: Long Papers)}, pages 1424--1433,
  Berlin, Germany, Aug. 2016. Association for Computational Linguistics.
\newblock \doi{10.18653/v1/P16-1135}.
\newblock URL \url{https://aclanthology.org/P16-1135}.

\bibitem[Hosseini et~al.(2021)Hosseini, Broniatowski, and
  Diab]{hosseini2021predicting}
P.~Hosseini, D.~A. Broniatowski, and M.~Diab.
\newblock Predicting directionality in causal relations in text.
\newblock \emph{arXiv preprint arXiv:2103.13606}, 2021.

\bibitem[Jin et~al.(2020)Jin, Wang, Luo, Huang, and Gu]{Jin2020IntersentenceAI}
X.~Jin, X.~Wang, X.~Luo, S.~Huang, and S.~Gu.
\newblock Inter-sentence and implicit causality extraction from chinese corpus.
\newblock \emph{Advances in Knowledge Discovery and Data Mining},
  12084:\penalty0 739 -- 751, 2020.

\bibitem[Khoo et~al.(1998)Khoo, Kornfilt, Oddy, and Myaeng]{khoo1998automatic}
C.~S. Khoo, J.~Kornfilt, R.~N. Oddy, and S.~H. Myaeng.
\newblock Automatic extraction of cause-effect information from newspaper text
  without knowledge-based inferencing.
\newblock \emph{Literary and Linguistic Computing}, 13\penalty0 (4):\penalty0
  177--186, 12 1998.
\newblock ISSN 0268-1145.
\newblock \doi{10.1093/llc/13.4.177}.
\newblock URL \url{https://doi.org/10.1093/llc/13.4.177}.

\bibitem[Khoo et~al.(2000)Khoo, Chan, and Niu]{khoo2}
C.~S.~G. Khoo, S.~Chan, and Y.~Niu.
\newblock Extracting causal knowledge from a medical database using graphical
  patterns.
\newblock In \emph{Proceedings of the 38th Annual Meeting on Association for
  Computational Linguistics}, ACL '00, page 336–343, USA, 2000. Association
  for Computational Linguistics.
\newblock \doi{10.3115/1075218.1075261}.
\newblock URL \url{https://doi.org/10.3115/1075218.1075261}.

\bibitem[Krippendorff(2011)]{krippendorff2011computing}
K.~Krippendorff.
\newblock Computing krippendorff's alpha-reliability.
\newblock 2011.

\bibitem[Laban et~al.(2021)Laban, Bandarkar, and Hearst]{Laban2021NewsHG}
P.~Laban, L.~Bandarkar, and M.~A. Hearst.
\newblock News headline grouping as a challenging nlu task.
\newblock In \emph{NAACL 2021}. Association for Computational Linguistics,
  2021.

\bibitem[Liu et~al.(2020)Liu, Peng, Li, Song, and
  Li]{event_evolution_social_streams}
Y.~Liu, H.~Peng, J.~Li, Y.~Song, and X.~Li.
\newblock Event detection and evolution in multi-lingual social streams.
\newblock \emph{Frontiers of Computer Science}, 14, 10 2020.
\newblock \doi{10.1007/s11704-019-8201-6}.

\bibitem[Prokhorenkova et~al.(2018)Prokhorenkova, Gusev, Vorobev, Dorogush, and
  Gulin]{catboost}
L.~Prokhorenkova, G.~Gusev, A.~Vorobev, A.~V. Dorogush, and A.~Gulin.
\newblock Catboost: Unbiased boosting with categorical features.
\newblock In \emph{Proceedings of the 32nd International Conference on Neural
  Information Processing Systems}, NIPS'18, page 6639–6649, Red Hook, NY,
  USA, 2018. Curran Associates Inc.

\bibitem[Radford et~al.(2019)Radford, Wu, Child, Luan, Amodei, and
  Sutskever]{radford2019language}
A.~Radford, J.~Wu, R.~Child, D.~Luan, D.~Amodei, and I.~Sutskever.
\newblock Language models are unsupervised multitask learners.
\newblock 2019.

\bibitem[Radinsky and Horvitz(2013)]{radinsky_2}
K.~Radinsky and E.~Horvitz.
\newblock Mining the web to predict future events.
\newblock In \emph{Proceedings of the Sixth ACM International Conference on Web
  Search and Data Mining}, WSDM '13, page 255–264, New York, NY, USA, 2013.
  Association for Computing Machinery.
\newblock ISBN 9781450318693.
\newblock \doi{10.1145/2433396.2433431}.
\newblock URL \url{https://doi.org/10.1145/2433396.2433431}.

\bibitem[Radinsky et~al.(2012)Radinsky, Davidovich, and
  Markovitch]{radinsky_prediction}
K.~Radinsky, S.~Davidovich, and S.~Markovitch.
\newblock Learning causality for news events prediction.
\newblock In \emph{Proceedings of the 21st International Conference on World
  Wide Web}, WWW '12, page 909–918, New York, NY, USA, 2012. Association for
  Computing Machinery.
\newblock ISBN 9781450312295.
\newblock \doi{10.1145/2187836.2187958}.
\newblock URL \url{https://doi.org/10.1145/2187836.2187958}.

\bibitem[Riaz and Girju(2013)]{riaz-girju-2013-toward}
M.~Riaz and R.~Girju.
\newblock Toward a better understanding of causality between verbal events:
  Extraction and analysis of the causal power of verb-verb associations.
\newblock In \emph{Proceedings of the {SIGDIAL} 2013 Conference}, pages 21--30,
  Metz, France, Aug. 2013. Association for Computational Linguistics.
\newblock URL \url{https://aclanthology.org/W13-4004}.

\bibitem[Ribeiro et~al.(2020)Ribeiro, Wu, Guestrin, and Singh]{checklist}
M.~T. Ribeiro, T.~Wu, C.~Guestrin, and S.~Singh.
\newblock Beyond accuracy: Behavioral testing of nlp models with checklist.
\newblock In \emph{Association for Computational Linguistics (ACL)}, 2020.

\bibitem[Roemmele et~al.(2011)Roemmele, Bejan, and Gordon]{roemmele2011choice}
M.~Roemmele, C.~A. Bejan, and A.~S. Gordon.
\newblock Choice of plausible alternatives: An evaluation of commonsense causal
  reasoning.
\newblock In \emph{2011 AAAI Spring Symposium Series}, 2011.
\newblock URL
  \url{https://people.ict.usc.edu/~gordon/publications/AAAI-SPRING11A.PDF}.

\bibitem[Shavrina et~al.(2020)Shavrina, Fenogenova, Emelyanov, Shevelev,
  Artemova, Malykh, Mikhailov, Tikhonova, Chertok, and
  Evlampiev]{shavrina2020russiansuperglue}
T.~Shavrina, A.~Fenogenova, A.~Emelyanov, D.~Shevelev, E.~Artemova, V.~Malykh,
  V.~Mikhailov, M.~Tikhonova, A.~Chertok, and A.~Evlampiev.
\newblock Russiansuperglue: A russian language understanding evaluation
  benchmark.
\newblock \emph{arXiv preprint arXiv:2010.15925}, 2020.

\bibitem[Wang et~al.(2019)Wang, Pruksachatkun, Nangia, Singh, Michael, Hill,
  Levy, and Bowman]{wang2019superglue}
A.~Wang, Y.~Pruksachatkun, N.~Nangia, A.~Singh, J.~Michael, F.~Hill, O.~Levy,
  and S.~R. Bowman.
\newblock Super{GLUE}: A stickier benchmark for general-purpose language
  understanding systems.
\newblock \emph{arXiv preprint 1905.00537}, 2019.

\bibitem[Xu et~al.(2020)Xu, Zuo, Liang, and Zuo]{review1}
J.~Xu, W.~Zuo, S.~Liang, and X.~Zuo.
\newblock A review of dataset and labeling methods for causality extraction.
\newblock In \emph{Proceedings of the 28th International Conference on
  Computational Linguistics}, pages 1519--1531, Barcelona, Spain (Online), Dec.
  2020. International Committee on Computational Linguistics.
\newblock \doi{10.18653/v1/2020.coling-main.133}.
\newblock URL \url{https://aclanthology.org/2020.coling-main.133}.

\bibitem[Yang et~al.(2009)Yang, Shi, and Wei]{event_evolution_graphs}
C.~C. Yang, X.~Shi, and C.-P. Wei.
\newblock Discovering event evolution graphs from news corpora.
\newblock \emph{Trans. Sys. Man Cyber. Part A}, 39\penalty0 (4):\penalty0
  850–863, July 2009.
\newblock ISSN 1083-4427.
\newblock \doi{10.1109/TSMCA.2009.2015885}.
\newblock URL \url{https://doi.org/10.1109/TSMCA.2009.2015885}.

\bibitem[Yang et~al.(2021)Yang, Han, and Poon]{review2}
J.~Yang, S.~C. Han, and J.~Poon.
\newblock A survey on extraction of causal relations from natural language
  text, 2021.

\end{thebibliography}

\section*{Checklist}

\begin{enumerate}

\item For all authors...
\begin{enumerate}
  \item Do the main claims made in the abstract and introduction accurately reflect the paper's contributions and scope?
    \answerYes{Section \ref{introduction}}
  \item Did you describe the limitations of your work?
    \answerYes{Section \ref{discussion}}
  \item Did you discuss any potential negative societal impacts of your work?
    \answerYes{Section \ref{introduction}, last paragraph}
  \item Have you read the ethics review guidelines and ensured that your paper conforms to them? 
    \answerYes{Section \ref{introduction}, last paragraph}
\end{enumerate}

\item If you are including theoretical results...
\begin{enumerate}
  \item Did you state the full set of assumptions of all theoretical results?
    \answerNA{}
	\item Did you include complete proofs of all theoretical results?
    \answerNA{}
\end{enumerate}

\item If you ran experiments (e.g. for benchmarks)...
\begin{enumerate}
  \item Did you include the code, data, and instructions needed to reproduce the main experimental results (either in the supplemental material or as a URL)?
    \answerYes{}
  \item Did you specify all the training details (e.g., data splits, hyperparameters, how they were chosen)?
    \answerYes{}
	\item Did you report error bars (e.g., with respect to the random seed after running experiments multiple times)?
    \answerYes{Table~\ref{simple_en_ru_results} and Table~\ref{full_en_ru_results}}
	\item Did you include the total amount of compute and the type of resources used (e.g., type of GPUs, internal cluster, or cloud provider)?
    \answerYes{Section~\ref{experiments}} 
\end{enumerate}

\item If you are using existing assets (e.g., code, data, models) or curating/releasing new assets...
\begin{enumerate}
  \item If your work uses existing assets, did you cite the creators?
    \answerYes{Footnotes in Section~\ref{sources}}
  \item Did you mention the license of the assets?
    \answerYes{Section~\ref{sources}}
  \item Did you include any new assets either in the supplemental material or as a URL?
    \answerYes{}
  \item Did you discuss whether and how consent was obtained from people whose data you're using/curating?
    \answerYes{}
  \item Did you discuss whether the data you are using/curating contains personally identifiable information or offensive content?
    \answerYes{Section \ref{introduction}, last paragraph}
\end{enumerate}

\item If you used crowdsourcing or conducted research with human subjects...
\begin{enumerate}
  \item Did you include the full text of instructions given to participants and screenshots, if applicable?
    \answerYes{Section~\ref{annotation}, Figure~\ref{fig:annotation_interface}}
  \item Did you describe any potential participant risks, with links to Institutional Review Board (IRB) approvals, if applicable?
    \answerNA{}
  \item Did you include the estimated hourly wage paid to participants and the total amount spent on participant compensation?
    \answerYes{Section~\ref{annotation}, Table~\ref{annot_stats}}
\end{enumerate}

\end{enumerate}

\end{document}